\documentclass[journal,12pt,draftclsnofoot,onecolumn]{IEEEtran}

\usepackage{verbatim}
\usepackage{amsmath,graphicx}
\usepackage{mathrsfs}
\usepackage{amsopn}
\usepackage{cite}
\usepackage{amsthm}
\usepackage{epsfig,psfrag}
\usepackage{amsfonts,amssymb}
\usepackage{subfigure}
\usepackage{algorithm} 
\usepackage{algorithmic} 
\usepackage{graphics}%
\usepackage{epsfig}
\usepackage{epstopdf}
\graphicspath{{Figures/}}
\usepackage{amsbsy}
\usepackage{tabularx}
\usepackage{amssymb} 
\usepackage{multirow} 
\usepackage{xcolor}
\usepackage{array}
\usepackage{arydshln}
\usepackage{color}
\usepackage{caption}
\usepackage{xcolor}
\usepackage{nicefrac}
\DeclareGraphicsExtensions{.pdf,.jpeg,.png}
\graphicspath{{./img/}}

\begin{document}

\title{A Fast Dictionary Learning Method for\\ 
	Coupled Feature Space Learning}

\author{Farshad~G.~Veshki 
	and Sergiy~A.~Vorobyov, {\it Fellow, IEEE} 
	\IEEEaftertitletext{\vspace{0\baselineskip}}
	\thanks{	
		F.~G.~Veshki and S.~A.~Vorobyov are with Aalto University, Dept. Signal Processing and Acoustics, FI-00076, AALTO, Finland. E-mails: farshad.ghorbaniveshki@aalto.fi, svor@ieee.org
		{\bf Corresponding author} is S.~A.~Vorobyov.}
	
}

\markboth{IEEE Signal Processing Letters}%
{Shell \MakeLowercase{\textit{et al.}}: Bare Demo of IEEEtran.cls for IEEE Journals}

\maketitle

\begin{abstract}
	In this letter, we propose a novel computationally efficient coupled dictionary learning method that enforces pairwise correlation between the atoms of dictionaries learned to represent the underlying feature spaces of two different representations of the same signals, e.g., representations in different modalities or representations of the same signals measured with different qualities. The jointly learned correlated feature spaces represented by coupled dictionaries are used in sparse representation based classification, recognition and reconstruction tasks. The presented experimental results show that the proposed coupled dictionary learning method has a significantly lower computational cost. Moreover, the visual presentation of jointly learned dictionaries shows that the pairwise correlations between the corresponding atoms are ensured.

\end{abstract}


\IEEEpeerreviewmaketitle

\section{Introduction}
\label{Introduction}
\IEEEPARstart{S}{parsity} and overcompleteness has been successfully used for diverse applications in signal processing over the last decade~\cite{Robus2009Face,Super2010Resolu,Kernel2012Sparse,Compressed2007Sensing}. The fact exploited is that signals can be compactly modelled using an overcomplete dictionary as a linear combination of only few \emph{atoms}. 

Formally, the basic {\it synthesis model} suggests that the signal $\boldsymbol{x}$ can be described as a linear combination of few atoms over an overcomplete dictionary $\boldsymbol{D}$, and the problem of seeking such sparse representation can be formulated as
$\underset{ {\boldsymbol{\alpha}} }{\mathrm{min}} \left \|{\boldsymbol{\alpha}} \right \|_0 \   
{\mathrm{s.t.}}\ \boldsymbol{x}\thickapprox\boldsymbol{D}\boldsymbol{\alpha}
$, where $\boldsymbol{\alpha}$ is the sparse vector of coefficients for atoms in the dictionary $\boldsymbol{D}$ and $\| \cdot \|_0$ denotes the operator that counts the number of non-zero entries in a vector. 

Many applications have benefited remarkably from using the above approach with learned overcomplete dictionary~\cite{Classification2010clustering,Sparsity2011based,Low2012Dose,Ana2013lysis}. Representative examples of dictionary learning algorithms include the K-SVD method~\cite{Aharon2006KSVD}, the method of optimal directions (MOD)~\cite{Engan1999MOD}, the online dictionary learning (OLD) method~\cite{Mairal2009ODl}, and their variants~\cite{rubinstein2008efficient,Discriminative,LabelConsistentKSVD}. ``Good'' dictionaries are expected to be highly adaptive to the observed signals and to lead to accurate sparse representations.

While the \emph{single dictionary} model has been extensively studied, there exists also a \emph{coupled dictionary} viewpoint to sparsity and overcompleteness, where a coupled dictionary is needed to represent the double feature space (e.g., focused and blurred image patches in image processing). The combination of learned coupled dictionary and sparse approximation is shown to be superior for representing double feature spaces~\cite{yang2012coupled,Sadasivan2016coupled, Wang2012coupled,Peleg2014Statistical,Selective2018sparse,Generalized2016Coupled,rui2016coupled,Coupled2013Dictionary}.

The coupled dictionary learning aims to find a pair of dictionaries $[ \boldsymbol{D}_1, \, \boldsymbol{D}_2 ]$ best representing two  subsets of $n$ training signals $\boldsymbol{X}_1=\left[[\boldsymbol{x}_1]_1,\cdots,[\boldsymbol{x}_1]_n\right]$ and $\boldsymbol{X}_2=[[\boldsymbol{x}_2]_1,\cdots,[\boldsymbol{x}_2]_n]$ in such a way that the atoms of $\boldsymbol{D}_1$ and $ \boldsymbol{D}_2$ are pairwise correlated, and if a linear combination of atoms of $\boldsymbol{D}_1$ models a signal in $\boldsymbol{X}_1$, the same linear combination of atoms of $\boldsymbol{D}_2$ also models the corresponding signal in $\boldsymbol{X}_2$. This can be insured by enforcing an identical sparse representation matrix $\boldsymbol{\Gamma}$ for both $\boldsymbol{X}_1$ and $\boldsymbol{X}_2$ while learning $ \boldsymbol{D}_1$ and $\boldsymbol{D}_2$. Then the coupled dictionary learning problem can be formulated as the following optimization problem~\cite{yang2012coupled}
\begin{equation}\label{eq1}
\begin{split}
\begin{aligned}	
&\underset{ {\boldsymbol{D}_1, \boldsymbol{D}_2, \boldsymbol{\Gamma}} }{\mathrm{min}} 
\left \| \boldsymbol{X}_1 - \boldsymbol{D}_1 \boldsymbol{\Gamma} \right \|_{\rm 2}^2 +\left \| \boldsymbol{X}_2 - \boldsymbol{D}_2 \boldsymbol{\Gamma} \right \|_{\rm 2}^2 \\
&{\mathrm{s.t.}}\ \left \| \boldsymbol{\gamma}_i^{\rm c} \right \|_0 \leqslant T_0,\ \left \| [\boldsymbol{d}_1]_t \right \|_{2}= 1, 
\left \| [\boldsymbol{d}_2]_t \right \|_{2}= 1,
\forall t,i	
\end{aligned}
\end{split}
\end{equation}
where $[\boldsymbol{d}_2]_t$ are the $t$-th dictionary atoms (columns) of $\boldsymbol{D}_1$ and $\boldsymbol{D}_2$, respectively, $T_0$ is the constraint value on sparsity, and $\|\cdot\|_2$ is the Euclidian norm of a vector. The notation $\boldsymbol{\gamma}_i^{\rm c}$ is used for $i$-th column of $\boldsymbol{\Gamma}$, to be distinct from the notation that later is used for the rows of the same matrix.

The methods in~\cite{yang2012coupled,Sadasivan2016coupled, Wang2012coupled,Selective2018sparse} address \eqref{eq1} to model the function between observation and latent feature spaces (e.g., noisy and clear data), so that they can recover the unknown higher quality signals from their available low quality versions. Inverse problems such as image superresolution~\cite{yang2012coupled,Selective2018sparse}, and speech signal bandwidth extension~\cite{Sadasivan2016coupled} are then examples of applications. For such methods, the corresponding dictionaries are expected to yield accurate sparse approximations. There are also methods that employ coupled dictionary learning techniques to solve problems such as cross-modal matching~\cite{Generalized2016Coupled}, cross-domain image recognition~\cite{Coupled2013Dictionary}, and multi-focus image fusion~\cite{rui2016coupled}, as examples of classification and recognition applications. In latter applications, the learned dictionaries are not required to provide accurate sparse recovery, but the objective is to learn the underlying feature spaces of $\boldsymbol{X}_1$ and $\boldsymbol{X}_2$, i.e., the coupled dictionary. 

A majority of existing coupled dictionary learning algorithms address \eqref{eq1} by learning two correlated feature spaces through burdensome complex procedures, while the computationally demanding nature of dictionary learning algorithms becomes more restrictive when we need to learn two dictionaries simultaneously. In this letter, we propose a fast coupled dictionary learning scheme that dramatically reduces the computational costs that brings it below the one by the K-SVD method even for a single dictionary.

\section{A NEW PROPOSED METHOD}
\label{The Proposed Method}

\begin{figure}[!htb]
	\centerline{\includegraphics[width=.75\textwidth]{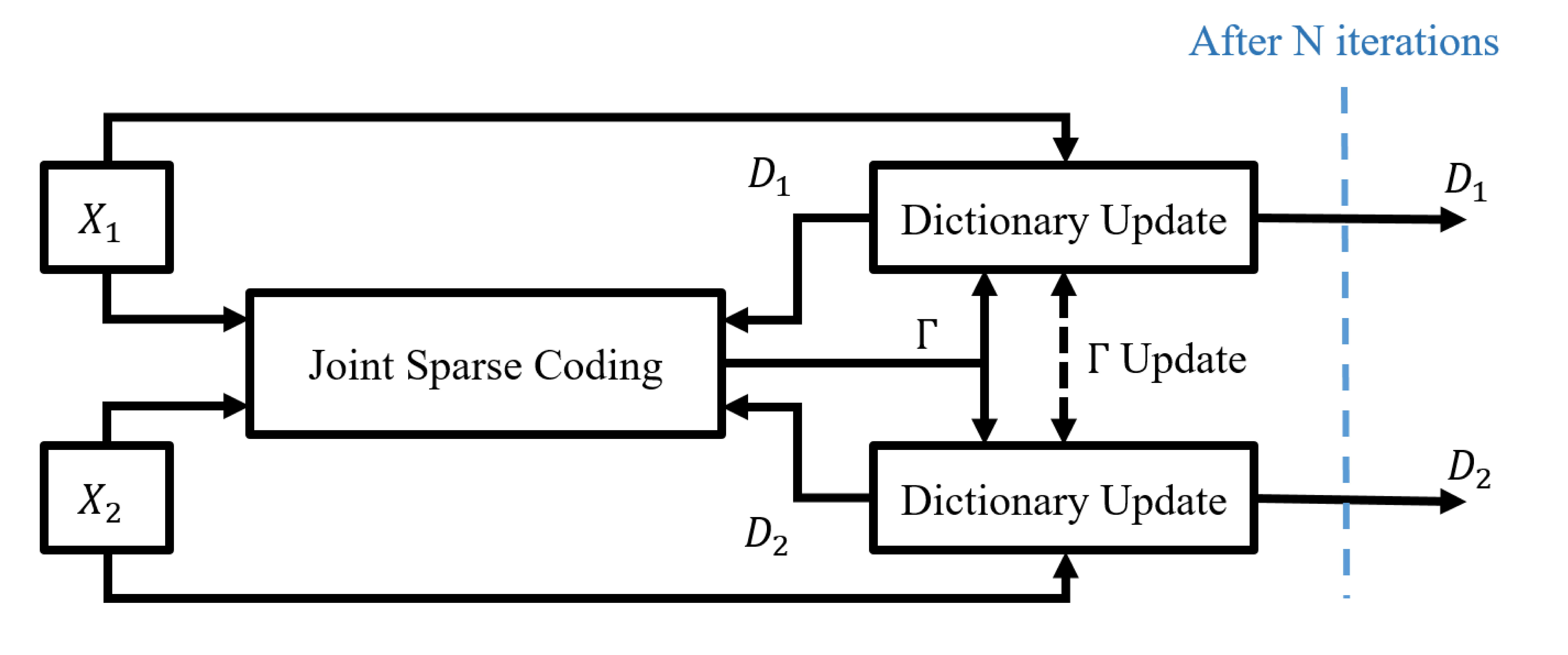}}
	\centering
	\caption{Block-diagram of the proposed coupled dictionary learning method.}
	\label{fig:diagram}
\end{figure}

The optimization variables in problem \eqref{eq1} can be split into two subsets, where one subset consists of the common sparse representation matrix $\boldsymbol{\Gamma}$, and the other includes the dictionaries $\boldsymbol{D}_{\rm 1}$ and $\boldsymbol{D}_{\rm 2}$. Then \eqref{eq1} can be addressed in alternating manner by iterating between two phases, where in the first phase $\boldsymbol{\Gamma}$ is optimized under the constraint $\left \| \boldsymbol{\gamma}_i^{\rm c} \right \|_0 \leqslant T_0$ -- a \emph{joint sparse coding} problem, and in the second phase $\boldsymbol{D}_{\rm 1}$ and $\boldsymbol{D}_{\rm 2}$ are optimized under the constraints $\ \left \| [\boldsymbol{d}_1]_t \right \|_{2}= 1$ and $ 
\left \| [\boldsymbol{d}_2]_t \right \|_{2}= 1$, respectively-- \emph{dictionary update} problems. The general procedure of the proposed coupled dictionary learning is summarized in the block-diagram presented in Fig.~\ref{fig:diagram}. In the dictionary update phase, after updating each atom, all nonzero coefficients of its corresponding row of $\boldsymbol{\Gamma}$ have to be updated. The dashed arrow in the block diagram indicates that in order to preserve the same sparse representation for both $\boldsymbol{D}_{\rm 1}$ and $\boldsymbol{D}_{\rm 2}$, the updates of $\boldsymbol{\Gamma}$ need to be performed jointly also during the dictionary update phase. Other operations, e.g., substituting unused atoms with better ones, are performed based on the common sparse representation matrix, thus the enforced atom-wise correlations in the joint sparse coding phase are preserved. The dictionaries can be initialized by any fixed basis overcomplete dictionary, e.g., discrete cosine transform (DCT) dictionary.
\vspace{-3mm}
\vspace{-5mm}
\subsection{Joint Sparse Coding}
\label{Joint Sparse Coding}
The joint sparse coding is the problem of finding optimal in least squares (LS) sense sparse representations of the joint dataset ${\boldsymbol{X}} \triangleq \left[ \boldsymbol{X}_1^T, \!\boldsymbol{X}_2^T\right]^T$ over the joint dictionary ${\boldsymbol{D}} \triangleq \left[ \boldsymbol{D}_1^T, \!\boldsymbol{D}_2^T\right]^T$, that is,
\begin{equation}\label{eq5}
\begin{split}
\begin{aligned}	
&\underset{\boldsymbol{\Gamma} }{\mathrm{min}} 
\left \| \boldsymbol{X} - \boldsymbol{D} \boldsymbol{\Gamma} \right \|_{\rm 2}^2 \quad {\mathrm{s.t.}}\ \quad \left \| \boldsymbol{\gamma}_i^{\rm c} \right \|_0 \leqslant T_0, \ 
\forall i.		
\end{aligned}
\end{split}
\end{equation}

Problem \eqref{eq5} is known to be NP-hard, but by replacing $\|~\cdot~\|_0$ with $l_1$-norm, it can be turned to a convex problem that is solvable by many existing methods. There are also sparse approximation methods known as matching pursuits (MP)~\cite{MatchingPursuits} which despite of not using explicit $l_1$-norm term, are proved to yield approximations for $l_1$-norm minimization problems~\cite{UnifiedOptimization}. 

Here, we use the orthogonal matching pursuit (OMP) method~\cite{OMP} to address the joint sparse coding problem. OMP is an iterative method that sequentially adds coefficients to the sparse representation vector in two steps. 

The first step is to find the best matching atom for the signal (or residual). The standard formulation for the matching problem is as follows
\begin{equation}\label{eq6}
\begin{split}
\begin{aligned}	
{\boldsymbol{d}_{best}}= \mathrm{arg}\underset{ {\boldsymbol{d}_ t}}{\mathrm{max}} 
\mid \boldsymbol{d}_t^T \boldsymbol{r} \mid
;\quad  \boldsymbol{d}_t \in \boldsymbol{D}
\end{aligned}
\end{split}
\end{equation}
where $\boldsymbol{d}_{best}$ denotes the best matching atom from the joint dictionary $\boldsymbol{D}$ for the joint residual $\boldsymbol{r} \triangleq \boldsymbol{x}_i - \boldsymbol{D}\boldsymbol{\gamma}_i^{\rm c}$ and $\boldsymbol{x}_i \in \boldsymbol{X}$. ``Matching'' is measured by the absolute value of correlation, i.e., $\mid\boldsymbol{d}_t^T \boldsymbol{r}\mid$. 

The second step is to calculate the coefficients for the atoms that are selected so far. This can be formulated as the following LS problem
\begin{equation}\label{eq7}
\begin{split}
\begin{aligned}
\underset{ \boldsymbol{\gamma}_i^{{\rm c}(m)} }{\mathrm{min}} 
\left \| \boldsymbol{r}^{(m)}  -  \boldsymbol{D}^{(m)} \boldsymbol{\gamma}_i^{{\rm c}(m)}  \right \|_{\rm 2}^2
\end{aligned}
\end{split}
\end{equation}
where $\boldsymbol{r}^{(m)}$ is the residual, $\boldsymbol{\gamma}_i^{{\rm c}(m)}$ is the sparse representation vector, and $\boldsymbol{D}^{(m)}$ is the subset of chosen atoms, all at $m$-th iteration. Problem \eqref{eq7} is equivalent to \eqref{eq1} optimized over $\boldsymbol{\Gamma}$ only, that is,
\begin{equation}\label{eq8}
\begin{split}
\begin{aligned}
\underset{ \boldsymbol{\gamma}_i^{{\rm c}(m)} }{\mathrm{min}} 
\left \| \boldsymbol{r}_1^{(m)}\!-\!\boldsymbol{D}_1^{(m)} \boldsymbol{\gamma}_i^{{\rm c}(m)}  \right \|_{\rm 2}^2\! +\! \left \| \boldsymbol{r}_2^{(m)}\!-\!\boldsymbol{D}_2^{(m)}\boldsymbol{\gamma}_i^{{\rm c}(m)} \right \|_{\rm 2}^2, \forall i
\end{aligned}
\end{split}
\end{equation}
where $\boldsymbol{r}_1$ and $\boldsymbol{r}_2$ are the residuals from $[\boldsymbol{x}_1]_i \in \boldsymbol{X}_1$ and $[\boldsymbol{x}_2]_i \in \boldsymbol{X}_2$, respectively. Thus, OMP approximates a common sparse representation matrix for $\boldsymbol{X}_1$ and $\boldsymbol{X}_2$ over $\boldsymbol{D}_1$ and $\boldsymbol{D}_2$, respectively. 

At the end of each iteration, the residuals need to be updated. 
The algorithm iterates until the remainder error which is calculated as the norm squared of the residuals $e = \left \| \boldsymbol{r}^{(m)} \right \|_{\rm 2}^2$ satisfies the error threshold $\epsilon$ or the number of coefficients reaches its limit $T_0$, i.e., the constraint $\left \| \boldsymbol{\gamma}_i^{\rm c} \right \|_0 \leqslant T_0$ is satisfied as equality. 
\vspace{-3mm}
\subsection{Dictionary Update}
\label{Dictionary Update}
For the common sparse representation $\boldsymbol{\Gamma}$, problem \eqref{eq1} needs to be solved then over the coupled dictionary $\boldsymbol{D}$. Since the objective function of \eqref{eq1} is separable with respect to the dictionaries ${\boldsymbol{D}_1}$ and ${\boldsymbol{D}_2}$, and different sets of constraints are applied to the atoms of ${\boldsymbol{D}_1}$ and ${\boldsymbol{D}_2}$, problem \eqref{eq1} can be split into two subproblems of finding updates for the dictionaries ${\boldsymbol{D}_1}$ and ${\boldsymbol{D}_2}$ separately, although the similarity of the sparse representations has to be maintained. Thus, we explain the proposed dictionary update for a single dictionary $\boldsymbol{D}_i$, $i=1,2$. 

The corresponding optimization problem is given as 
\begin{equation}\label{eq9}
\begin{split}
\begin{aligned}	
\ \boldsymbol{D}_i = \mathrm{arg}\underset{\boldsymbol{D}_i} {\mathrm{min}} 
\left \| \boldsymbol{X}_i- \sum\limits_{t}^{} [\boldsymbol{d}_i]_t \boldsymbol{\gamma}_t^{\rm r} \right \|_{\rm F}^2
\end{aligned}
\end{split}
\end{equation}
subject to the constraints in \eqref{eq1} applicable to corresponding atoms. Here $\boldsymbol{\gamma}_t^{\rm r} $ is the $t$-th row of $\boldsymbol{\Gamma}$. Note that in \eqref{eq9}, we rewrite the product $\boldsymbol{D}\boldsymbol{\Gamma}$ as the sum of vector outer products $[\boldsymbol{d}_i]_t \boldsymbol{\gamma}_t^{\rm r}$. After such modification, it appears that each atom can be updated disjoint from the others. Thus, to update the atom $[\boldsymbol{d}_i]_t$, we fix the remaining atoms, and rewrite optimization problem \eqref{eq9} as \vspace{-1mm}
\begin{equation}\label{eq10}
\begin{split}
\begin{aligned}		
{[\boldsymbol{d}_i]_t} = \mathrm{arg} \underset{[\boldsymbol{d}_i]_t}{\mathrm{min}} \left \|\! \left(\!\boldsymbol{X}_i - \sum\limits_{s\neq t} [\boldsymbol{d}_i]_s \boldsymbol{\gamma}_s^{\rm r}  \!\right) - [\boldsymbol{d}_i]_t \boldsymbol{\gamma}_t^{\rm r} \right \|_{\rm F}^2.
\end{aligned}
\end{split}
\end{equation}

Columns of $\boldsymbol{X}_i\! -\! \sum_{s\neq t} [\boldsymbol{d}_i]_s \boldsymbol{\gamma}_s^{\rm r} $ that correspond to zero entries of $\boldsymbol{\gamma}_t^{\rm r} $ can be ignored. Thus, we define the vector $\boldsymbol{\omega_t}$ representing the subset of indices where $\boldsymbol{\gamma}_t^{\rm r} \neq 0$, that is, $\boldsymbol{\omega_t}\!=\!\!\{i| [\boldsymbol{\gamma}_t^{\rm r} ]_i\!\neq\!0\}$.
Then the error matrix $[\boldsymbol{E}_i]_t$ is formed as \vspace{-1mm}
\vspace{-1mm}
\begin{equation}\label{eq12}
\begin{split}
\begin{aligned}	
{[\boldsymbol{E}_i]_t} \triangleq \left[\boldsymbol{X}_i - \sum_{s\neq t} [\boldsymbol{d}_i]_s \boldsymbol{\gamma}_t^{\rm r} \right]_{\boldsymbol{\omega_t}}.
\end{aligned}
\end{split}
\end{equation}

Then optimization problem~\eqref{eq10} can be further rewritten as the following simple rank-1 LS approximation problem  
\begin{equation}\label{eq13}
\begin{matrix}
{[\boldsymbol{d}_i]_t} = \mathrm{arg} \underset{ [\boldsymbol{d}_i]_t }{\mathrm{min}} \left \| [\boldsymbol{E}_i]_t - [\boldsymbol{d}_i]_t [\boldsymbol{\gamma}_t^{\rm r} ]_{\boldsymbol{\omega_t}} \right \|_{\rm F}^2 \quad 
\end{matrix}
\end{equation}
where $[\boldsymbol{\gamma}_t^{\rm r}]_{\boldsymbol{\omega_t}}$ contains nonzero entries of $\boldsymbol{\gamma}_t^{\rm r}$. There is no sparsity constraint in LS problem~\eqref{eq13}, thus, it can be easily solved as ${\boldsymbol{{d}}}_t \!=\! \boldsymbol{E}_t[\boldsymbol{\gamma}_t^{\rm r} ]_{\boldsymbol{\omega_t}}^T /\| [\boldsymbol{\gamma}_t^{\rm r} ]_{\boldsymbol{\omega_t}}\|_2^2 $. The normalization term $\| [\boldsymbol{\gamma}_t^{\rm r} ]_{\boldsymbol{\omega_t}}\|_2^2$ can be dropped, since we need to normalize the $l_2$-norm of each atom to one anyway. Then the atom update rule is
\begin{equation}\label{eq14}
\begin{matrix}
{[\boldsymbol{d}_i]_t} = [\boldsymbol{E}_i]_t[\boldsymbol{\gamma}_t^{\rm r} ]_{\boldsymbol{\omega_t}}^T.
\end{matrix}
\end{equation}

If $\boldsymbol{\omega_t}$ is empty, $[\boldsymbol{d}_i]_t$ is updated as the column-wise average of error matrix $[\boldsymbol{E}_i]_t=\boldsymbol{X}_i-\boldsymbol{D}_i \boldsymbol{\Gamma}$. To avoid the scale ambiguity in sparse approximation, the updated atoms are then normalized. 

After updating $[\boldsymbol{d}_i]_t$, we need to update $[\boldsymbol{\gamma}_t^{\rm r} ]_{\boldsymbol{\omega_t}}$ accordingly. Since $[\boldsymbol{d}_i]_t$ is a unit vector, the solution of \eqref{eq13}, this time over $[\boldsymbol{\gamma}_t^{\rm r} ]_{\boldsymbol{\omega_t}}$, can be efficiently found as
$[\boldsymbol{\gamma}_t^{\rm r} ]_{\boldsymbol{\omega_t}} = [\boldsymbol{d}_i]_t^T[\boldsymbol{E}_i]_t$. However, this solution is different for each feature space, i.e., $i=1$ and $i=2$. Thus, the optimal common nonzero coefficients can be found for the joint atom $\boldsymbol{d}_t =\left[[\boldsymbol{d}_1]_t^T, [\boldsymbol{d}_2]_t^T\right]^T$ and joint error matrix $\boldsymbol{E}_t = \left[[\boldsymbol{E}_1]_t^T, [\boldsymbol{E}_2]_t^T\right]^T$, as
\begin{equation}\label{eq15}
\begin{matrix}
[\boldsymbol{\gamma}_t^{\rm r} ]_{\boldsymbol{\omega_t}} = \frac{1}{2}\boldsymbol{d}_t^T\boldsymbol{E}_t.
\end{matrix}
\end{equation} 

The complexity orders of \eqref{eq14} and \eqref{eq15} are both $\boldsymbol{O}(mn)$, which is much smaller than that of singular value decomposition (SVD) in \cite{Aharon2006KSVD} with complexity order of $\boldsymbol{O}(\mathrm{max}(m,n)^2\times\mathrm{min}(m,n))$.

\subsection{Maximum Number of Nonzero Coefficients}
\label{Coefs}
In each iteration, the majority of the existing two-phased alternating dictionary learning methods (including~\cite{Aharon2006KSVD,Engan1999MOD,Mairal2009ODl,rubinstein2008efficient,Discriminative,LabelConsistentKSVD}) first find $\boldsymbol{\Gamma}$ over $\boldsymbol{D}$, then update the atoms to reduce the error $\|\boldsymbol{X}-\boldsymbol{D}\boldsymbol{\Gamma}\|_2^2 $ in order to have a sparser $\boldsymbol{\Gamma}$ in the next iteration. That means that $\boldsymbol{\Gamma}$ is not sparse enough at the beginning. This backward approach imposes unnecessary extra computational costs, since a larger number of nonzero entries in sparse representation matrix leads to higher computational costs in both sparse coding and dictionary update phases. 

Another drawback of this backward approach is that it reduces the effectiveness of the dictionary update phase. Each atom $\boldsymbol{d}_t$ is updated according to the error matrix $\boldsymbol{E}_t$, which represents a potential amount of error that the atom update can compensate for in the total approximation error. When the dictionary is not learned to yield sparse enough approximations, the backward approach adds more coefficients to the sparse representations to minimize the approximation error, which leads to smaller entries for $\boldsymbol{E}_t$, thus reducing the learning potential for updating $\boldsymbol{d}_t$.

These issues can be easily addressed. Instead of setting the maximum number of nonzero coefficients as a constant number, we can gradually increase it. As a result, the first iterations become computationally cheap and the dictionary update phase becomes more effective. For example, we can form a vector of a size equal to the number of update cycles of dictionary learning algorithm, and set its values as equally spaced numbers between a minimum (e.g., 1) and the maximum number of nonzero coefficients. This simple change significantly reduces the computational cost without sacrificing the performance, even slightly.

\subsection{Summary of the Algorithm}
\label{Summary}
The overall algorithm for coupled dictionary learning can be then summarized as in Algorithm~\ref{alg:coupled}, where lines 3 to 11 represent the sparse coding phase, and lines 12 to 18 represent the dictionary update phase.

\begin{algorithm}[htb]    
	\renewcommand{\algorithmicrequire}{\textbf{Input:}}
	\renewcommand{\algorithmicensure}{\textbf{Output:}} 
	\caption{Coupled Dictionary Learning.}   
	\label{alg:coupled}
	\begin{algorithmic}[1]     
		\REQUIRE Two training datasets of $N$ signals $\boldsymbol{X}_{\rm 1}$ and $\boldsymbol{X}_{\rm 2}$, and $\boldsymbol{D}_0=$ DCT dictionary.
		\STATE  \textbf{Initialization}: Set $\boldsymbol{D}_1 :=\boldsymbol{D}_0$, $\boldsymbol{D}_2 := \boldsymbol{D}_0$.\\
		Number of update cycles $:= N$.\\
		$\boldsymbol{maxNum}$ = A sequence of $N$ equally spaced numbers between 1 and the maximum number of nonzero coefficients.\\
		
		\STATE \textbf{for} $k=1 \cdots N$ \textbf{do} 
		\STATE \quad \textbf{for} $i=1 \cdots n$ \textbf{do} \\
		\STATE \qquad Set $\boldsymbol{r} = \left[ [\boldsymbol{x}_1]_i^T, \ [\boldsymbol{x}_2]_i^T \right]^T$;\\
		\qquad$m \gets 1$;
		\STATE \qquad \textbf{while} $e\!>\!\epsilon$ and 
		$m \leqslant \boldsymbol{maxNum}(k)$
		\STATE  \qquad \quad Find $\boldsymbol{d}_{best}$ by solving \eqref{eq6};\\
		\STATE  \qquad \quad Find $\boldsymbol{\gamma}_t^{{\rm c}(m)}$ by solving \eqref{eq7};\\
		\STATE  \qquad \quad Update $\boldsymbol{r}^{(m)}= \boldsymbol{x}_i \!-\! \boldsymbol{D}\boldsymbol{\gamma}_t^{{\rm c}}$;\\
		\STATE  \qquad \quad Update $e = \left \| \boldsymbol{r}^{(m)} \right \|_{\rm 2}^2$;\\
				$\quad \qquad m \gets m+1$;			
		\STATE \qquad \textbf{end while}
		\STATE \quad \textbf{end for}
		\STATE \quad \textbf{for} $t=1 \cdots$ number of atoms \textbf{do} \\
		\STATE  \qquad Find $\boldsymbol{\omega_t} = \{i| [\boldsymbol{\gamma}_t^{{\rm r}}]_i \neq 0\}$; \\
		\STATE  \qquad Find $[\boldsymbol{E}_1]_t$ and $[\boldsymbol{E}_2]_t$ for $[\boldsymbol{d}_1]_t$ and $[\boldsymbol{d}_2]_t$  using \eqref{eq12}; \\
		\STATE  \qquad Update $[\boldsymbol{d}_1]_t$ and $[\boldsymbol{d}_2]_t$ using \eqref{eq14}; \\
		\STATE  \qquad Normalize the atoms:\\
		\qquad $[\boldsymbol{d}_1]_t=\nicefrac{[\boldsymbol{d}_1]_t}{\|[\boldsymbol{d}_1]_t\|_2}$ and $[\boldsymbol{d}_2]_t=\nicefrac{[\boldsymbol{d}_2]_t}{\|[\boldsymbol{d}_2]_t\|_2}$;\\
		\STATE  \qquad Update $\boldsymbol{\gamma}_t^{{\rm r}}$ using \eqref{eq15}; \\	 
		\STATE \quad \textbf{end for}
		\STATE \textbf{end for}
		\ENSURE  The pairwise correlated dictionaries $ \boldsymbol{D}_{\rm 1}$ and $ \boldsymbol{D}_{\rm 2}$.\
	\end{algorithmic}
\end{algorithm}

\section{EXPERIMENTAL RESULTS}
\label{RESULTS}
\begin{figure}[htb]
	\begin{center}
		\begin{minipage}{0.45\linewidth}
			\centerline{$(a)$}
		\end{minipage}%
		\begin{minipage}{0.45\linewidth}
			\centerline{$(b)$}
		\end{minipage}%
		
		\vfill
		\begin{minipage}{0.45\linewidth}
			\centerline{\includegraphics[width=5.5cm,height=5.5cm]{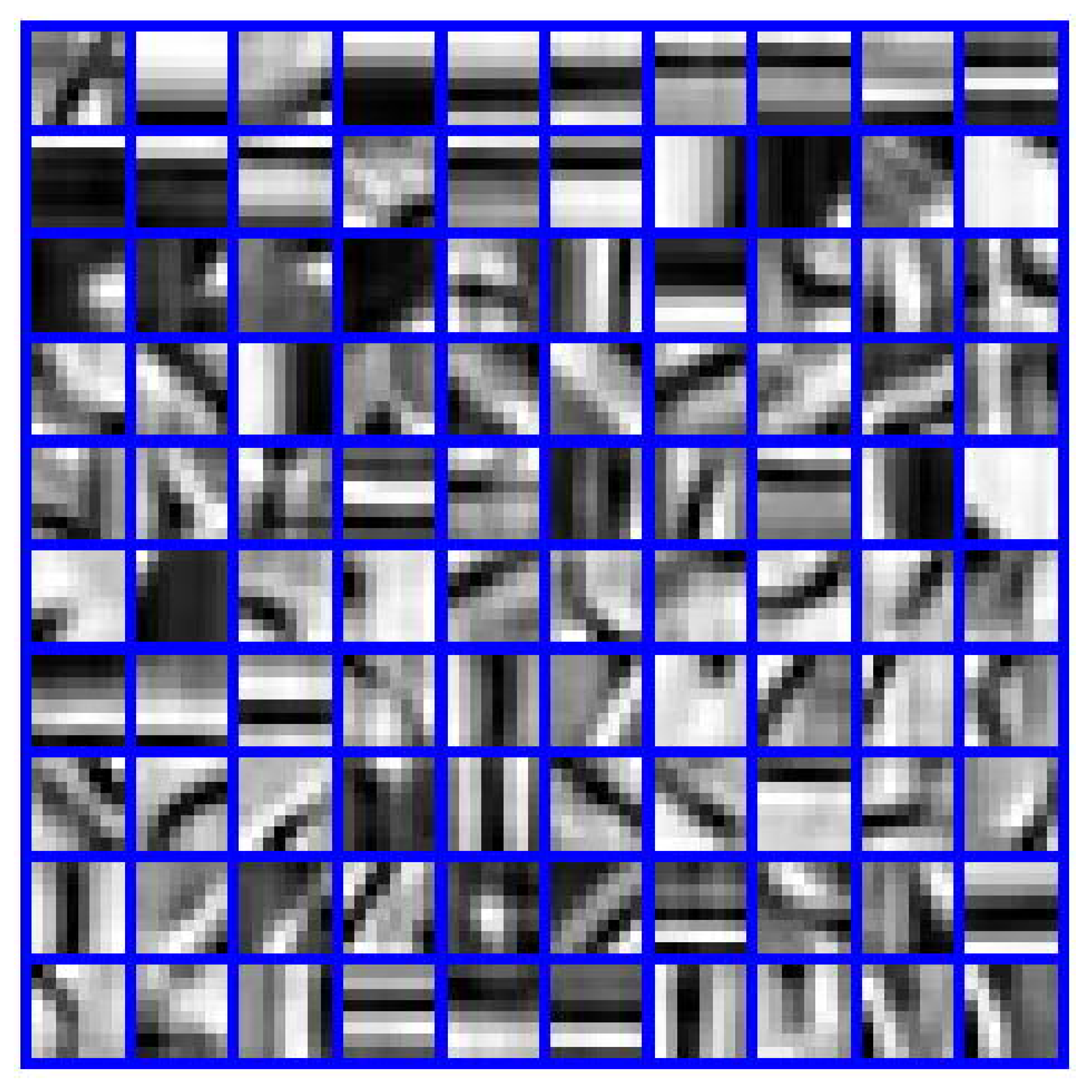}}
		\end{minipage}%
		\begin{minipage}{0.45\linewidth}
			\centerline{\includegraphics[width=5.5cm,height=5.5cm]{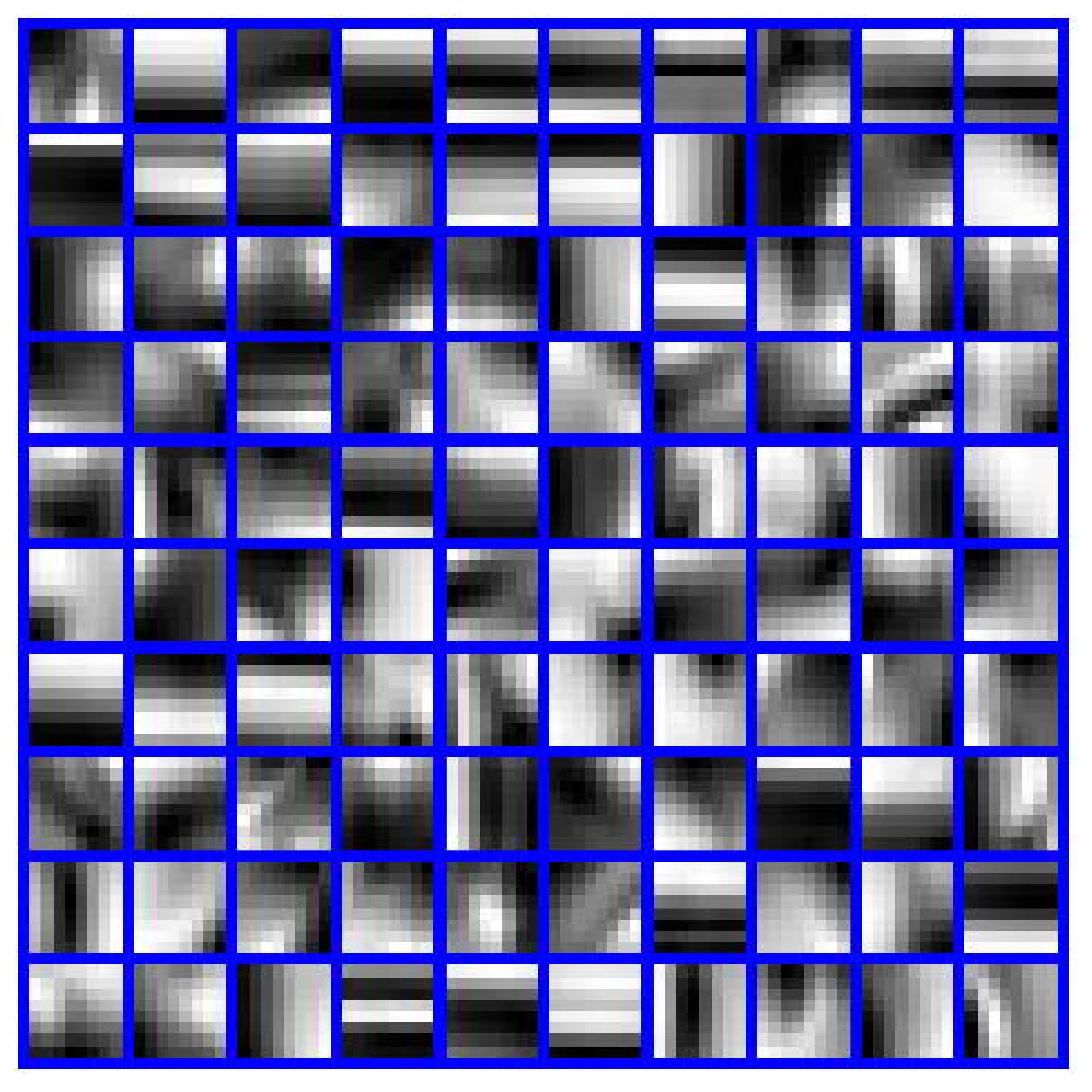}}
		\end{minipage}%
		\vfill
		\begin{minipage}{0.45\linewidth}
			\centerline{$(c)$}
		\end{minipage}%
		\begin{minipage}{0.45\linewidth}
			\centerline{$(d)$}
		\end{minipage}%
		
		\vfill
		\begin{minipage}{0.45\linewidth}
			\centerline{\includegraphics[width=5.5cm,height=5.5cm]{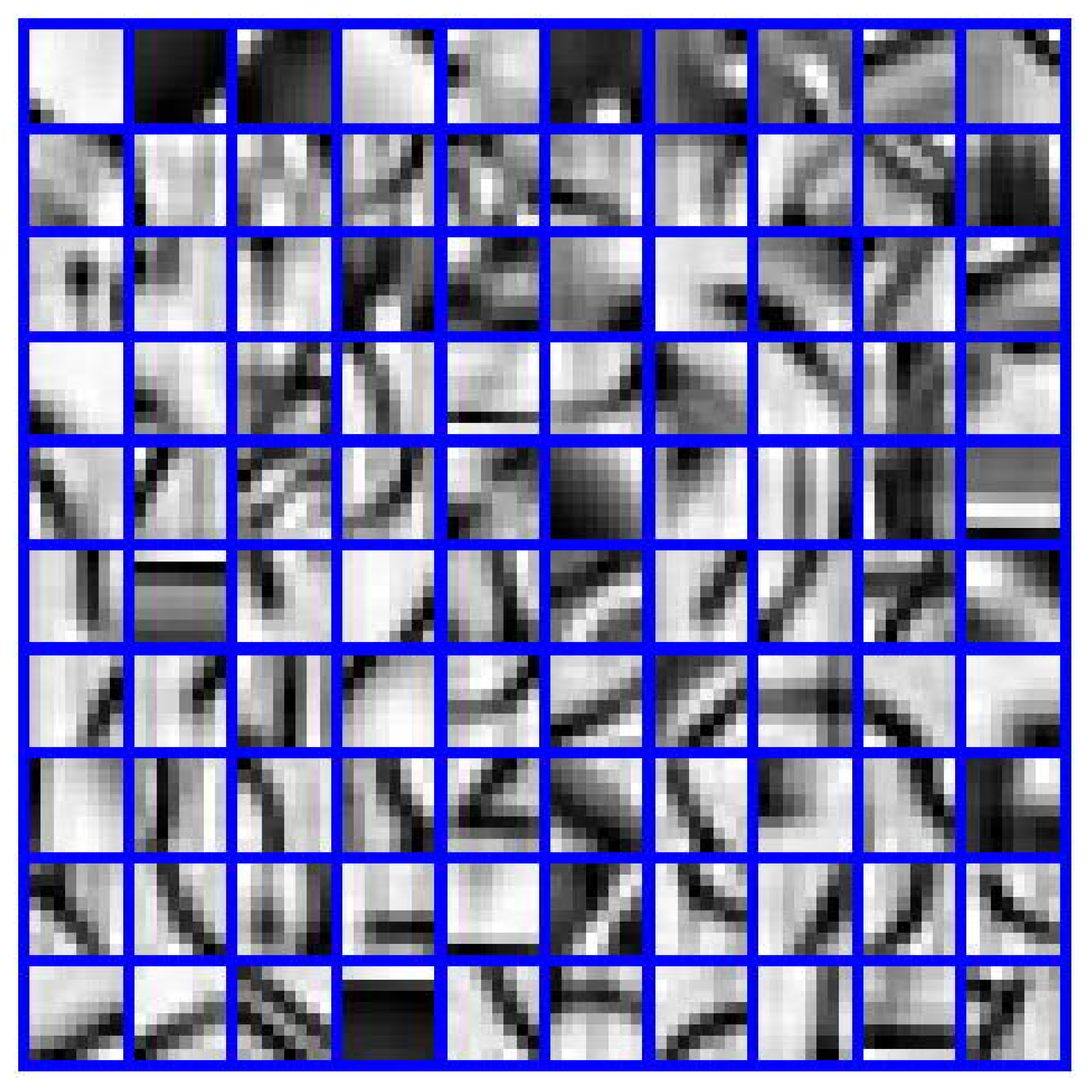}}
		\end{minipage}%
		\begin{minipage}{0.45\linewidth}
			\centerline{\includegraphics[width=5.5cm,height=5.5cm]{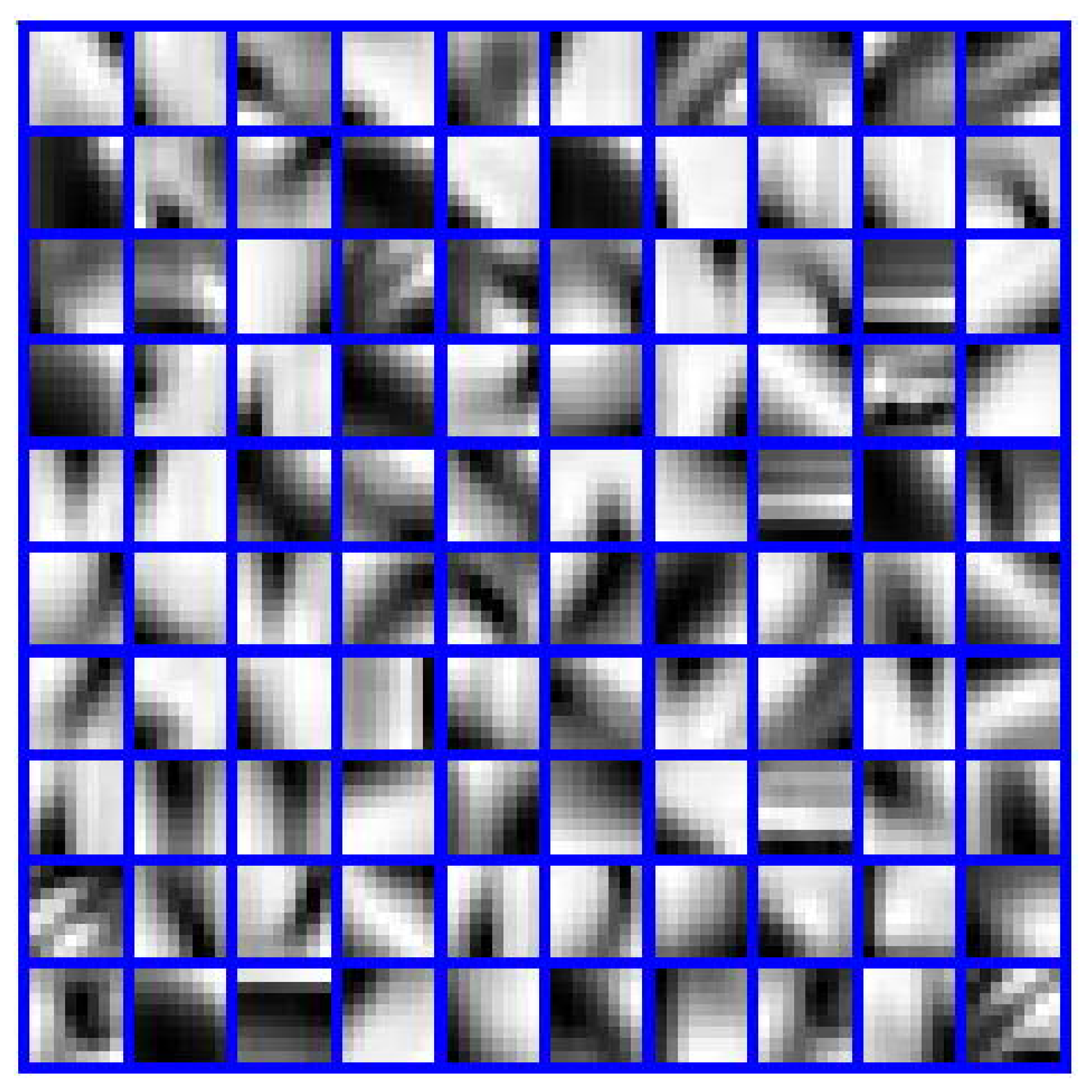}}
		\end{minipage}%
		\centering{
			\caption{Visual comparison between coupled learned dictionaries: (a) $\boldsymbol{D}_{\rm 1}$ and (b) $\boldsymbol{D}_{\rm 2}$, and separately learned dictionaries: (c) $\boldsymbol{D}_{\rm 1}$ and (d) $\boldsymbol{D}_{\rm 2}$.}	
			\label{fig:CoupledDict}}
	\end{center}
\end{figure}

In this section, we first demonstrate that the proposed coupled dictionary learning method is able to provide the desired pairwise correlation between the atoms of two jointly learned dictionaries. As an example experiment, we generate two subsets of 20,000 focused and blurred $8\times8$ grayscale image patches taken from Lytro image dataset~\cite{lytro}, and use them as $\boldsymbol{X}_1$ (focused data) and $\boldsymbol{X}_2$ (blurred data), where the patches (signals) in $\boldsymbol{X}_2$ are blurred versions of their corresponding focused patches in $\boldsymbol{X}_1$. The columns of $\boldsymbol{X}_1$ and $\boldsymbol{X}_2$ are vectorized image patches. We apply our method to the double feature space and learn the correlated dictionaries $\boldsymbol{D}_1$ and $\boldsymbol{D}_2$ (see Figs.~\ref{fig:CoupledDict}.(a) and (b)), then we visually compare it to the case where $\boldsymbol{D}_1$ and $\boldsymbol{D}_2$ are learned separately from the same feature spaces (see Figs.~\ref{fig:CoupledDict}.(c) and (d)).

From the visual representations of atoms if Fig.~\ref{fig:CoupledDict}, the pairwise correlations can be observed only between the atoms of dictionaries learned by the proposed coupled dictionary learning method. Those correlations are obtained by enforcing identical sparse representations through the proposed method and ensure that $\boldsymbol{D}_1$ and $\boldsymbol{D}_2$ represent corresponding features from the focused and blurred feature spaces.

 Next, we compare our proposed dictionary learning method to the K-SVD and ODL methods, in terms of runtime, obtained number of nonzero coefficients, and average learning error $\sqrt{\sum_{i=1}^{n}(\boldsymbol{x}_i-\boldsymbol{D}\boldsymbol{\gamma}_i^{\rm c})^2}/n$, for learning a dictionary from a single feature space.\footnote{Note that the proposed method is applicable without any change to a single dictionary learning as well.} The experiment is performed on a PC running an Intel(R) Xeon(R) 3.40GHz~CPU. The learning dataset includes 10,000 mean centred grayscale image patches with the size of $8\times8$, taken from Lytro image dataset. The tolerance error is set as $\epsilon = 4$, and the maximum number of nonzero coefficients is set to 32 (half of the size of vectorized patches). We run the K-SVD method for 16, the proposed algorithm for 32, and the ODL method for 256 dictionary learning cycles. The numbers of learning cycles are chosen with regards to the computational costs of the iterations of the algorithms, in a way that the ultimate runtimes are almost the same, so we can compare the results.
\vspace{-2mm}
\begin{figure}[htb]
	\begin{center}
		\begin{minipage}{1\linewidth}
			\centerline{$(a)$}
		\end{minipage}%
		
		\vfill
			\centerline{\includegraphics[width=9cm,height=4cm]{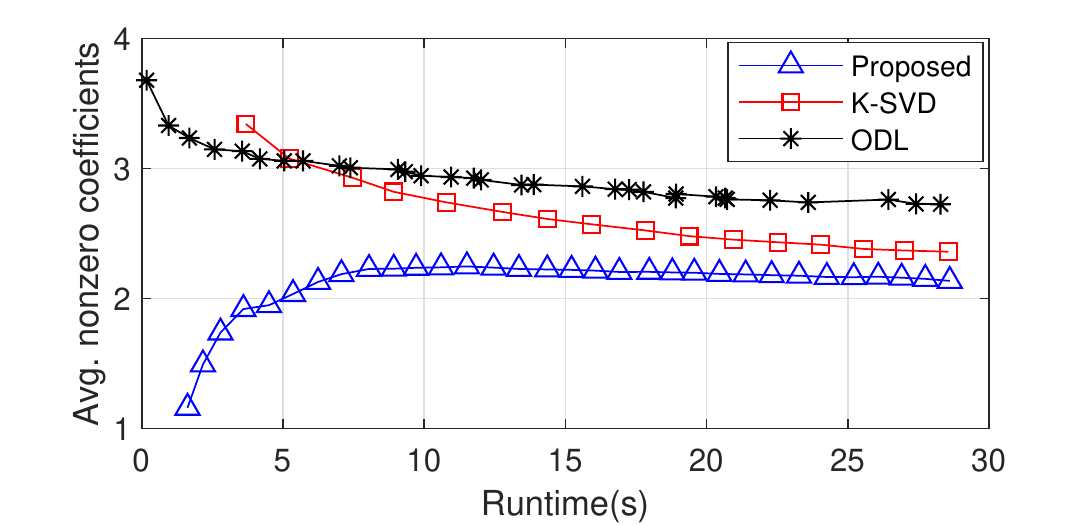}}
		\vfill
		\begin{minipage}{1\linewidth}
			\centerline{$(b)$}
		\end{minipage}%
		\vfill
			\centerline{\includegraphics[width=9cm,height=4cm]{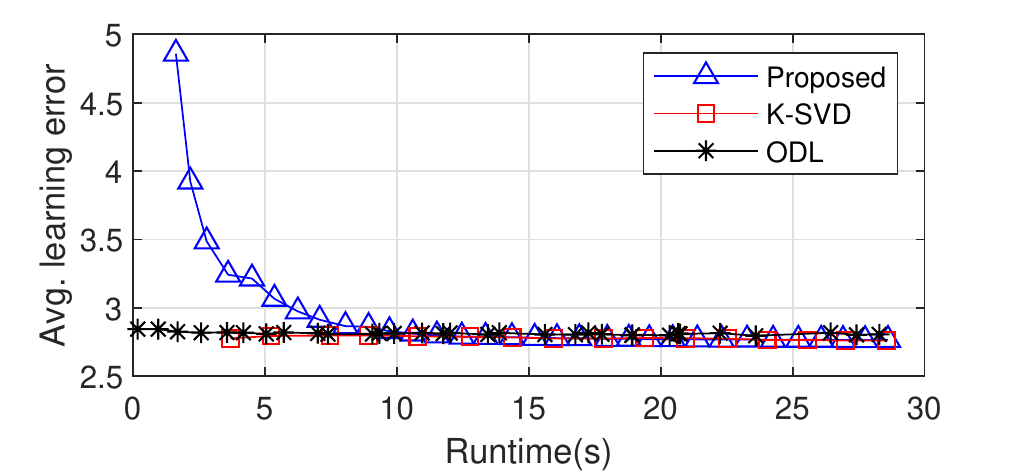}}
		\centering{
			\caption{The results for (a) the average number of nonzero coefficients, and (b) the average learning error versus the runtime. For the K-SVD and proposed methods, the markers indicate each iteration. For the ODL method, the markers show each 8 iterations.}	
			\label{fig:DicCompare}}
	\end{center}
\end{figure}

From Fig.~\ref{fig:DicCompare}(a), it can be observed that the dictionary learned by the proposed method yields significantly sparser representations in a much shorter time, comparing to those learned by the other methods. To explain this result, we visualize the changes in average learning error in Fig.~\ref{fig:DicCompare}(b).
As explained in Subsection~\ref{Coefs}, in the proposed method, we increase the maximum number of nonzero coefficients gradually. As a result, in the first iterations, the average error is high, however those iterations are faster. In this experiment, when the K-SVD method finishes its fifth iteration, the proposed algorithm has iterated 12 times. After about 11 seconds (13 iterations), the proposed method reaches the same average error and obtains a sparsity level which the K-SVD method achieves in about 28 seconds (16 iterations). 

\section{Conclusion}
\label{Conclusion}
A novel fast coupled dictionary learning algorithm that enforces common sparse approximations for double feature spaces and learns correlated pairs of atoms representing corresponding features from different feature spaces has been developed. The proposed dictionary learning method reduces dramatically the computational cost, which is important for computationally costly tasks such as coupled dictionary learning. The proposed method can be straightforwardly extended to find joint dictionaries for more than two feature spaces.

%

\






\end{document}